\documentclass[lettersize,journal]{IEEEtran}
\usepackage{amsmath,amsfonts}
\usepackage{algorithmic}
\usepackage{array}
\usepackage[caption=false,font=normalsize,labelfont=sf,textfont=sf]{subfig}
\usepackage{textcomp}
\usepackage{stfloats}
\usepackage{url}
\usepackage{verbatim}
\usepackage{graphicx}
\def\BibTeX{{\rm B\kern-.05em{\sc i\kern-.025em b}\kern-.08em
    T\kern-.1667em\lower.7ex\hbox{E}\kern-.125emX}}
\usepackage{balance}
\usepackage{subfiles}
\usepackage{booktabs}
\usepackage{multirow}
\usepackage{cite}
\usepackage[table]{xcolor}
\usepackage{makecell}
\usepackage{bm}

\DeclareSubrefFormat{parens}{#1(#2)}

\begin{document}
\title{PolarBEVDet: Exploring Polar Representation for Multi-View 3D Object Detection in Bird's-Eye-View}
\author{Zichen~Yu,
        Quanli~Liu,~\IEEEmembership{Member,~IEEE},
        Wei~Wang,~\IEEEmembership{Senior Member,~IEEE}, \\
        Liyong~Zhang,~\IEEEmembership{Member,~IEEE},
        and Xiaoguang~Zhao
        \IEEEcompsocitemizethanks{
            \IEEEcompsocthanksitem~Zichen Yu, Quanli Liu, Wei Wang and Liyong Zhang are with the School of Control Science and Engineering, 
            Dalian University of Technology, Dalian 116024, China, and also with the Dalian Rail Transmit Intelligent Control and 
            Intelligent Operation Technology Innovation Center, Dalian 116024, China (e-mail: yuzichen@mail.dlut.edu.cn; liuql@dlut.edu.cn; 
            wangwei@dlut.edu.cn; zhly@dlut.edu.cn).
            \IEEEcompsocthanksitem~Xiaoguang Zhao is with the Dalian Rail Transmit Intelligent Control and Intelligent Operation Technology 
            Innovation Center, Dalian 116024, China, and also with the Dalian Seasky Automation Co., Ltd, Dalian 116024, China (e-mail: xiaoguang.zhao@dlssa.com).
            \IEEEcompsocthanksitem~This work was supported in part by the National Natural Science Foundation of China under Grant 62373077
            and in part by the Key Field Innovation Team Support Plan of Dalian, China, under Grant No.2021RT02.
            \textit{(Corresponding author: Quanli Liu).} 
        }
}

\maketitle

\begin{abstract}
    Multi-view 3D object detection built upon the Lift-Splat-Shoot~(LSS) mechanism provides an economical and deployment-friendly solution 
    for autonomous driving. However, all existing LSS-based methods rely on transforming multi-view image features 
    into a Cartesian Bird's-Eye-View (BEV) representation, neglecting the non-uniformity of the image information 
    distribution, which results in loss information in the near and computational redundancy in the far. 
    Furthermore, these methods struggle to exploit view symmetry, increasing the difficulty of representation learning. 
    In this paper, to fundamentally remove these limitations, we propose to replace the Cartesian BEV representation 
    with the polar BEV representation, which naturally adapts to the image information distribution and effortlessly 
    preserves view symmetry by regular convolution. To achieve this, we elaborately tailor three modules: a polar 
    view transformer to generate the polar BEV representation, a polar temporal fusion module for fusing historical 
    polar BEV features and a polar detection head to predict the polar-parameterized representation of the object. 
    In addition, we design a 2D auxiliary detection head and a spatial attention enhancement module to improve the 
    quality of feature extraction in perspective view and BEV, respectively. Finally, we integrate the 
    above improvements into a novel multi-view 3D object detector, PolarBEVDet. Experiments on nuScenes show that 
    PolarBEVDet achieves superior performance, and the polar BEV representation can be seamlessly substituted into 
    different LSS-based detectors with consistent performance improvement. The code is available at 
    https://github.com/Yzichen/PolarBEVDet.git.
\end{abstract}

\section{Introduction}
\label{sec:intro}
\IEEEPARstart{M}{ulti-view} 3D object detection is a promising technology in the field of autonomous driving 
due to its low-cost deployment and rich semantic information~\cite{wang2024bevrefiner, shu20233dppe, Li2024Delving}. 
Early attempts~\cite{wangfos3d, park2021pseudo, Reading2021caddn, wang2022pgd, Sheng2023PDR} approach this task 
primarily from the perspective of monocular 3D object detection, where 3D object detection is first performed 
for each view, and then the predictions from all views are fused through post-processing. 
While feasible, this detection scheme ignores information across views, resulting in sub-optimal performance.

\begin{figure}[t]
    \centering
    \subfloat[Cartesian BEV]{
      \includegraphics[width=0.45\linewidth]{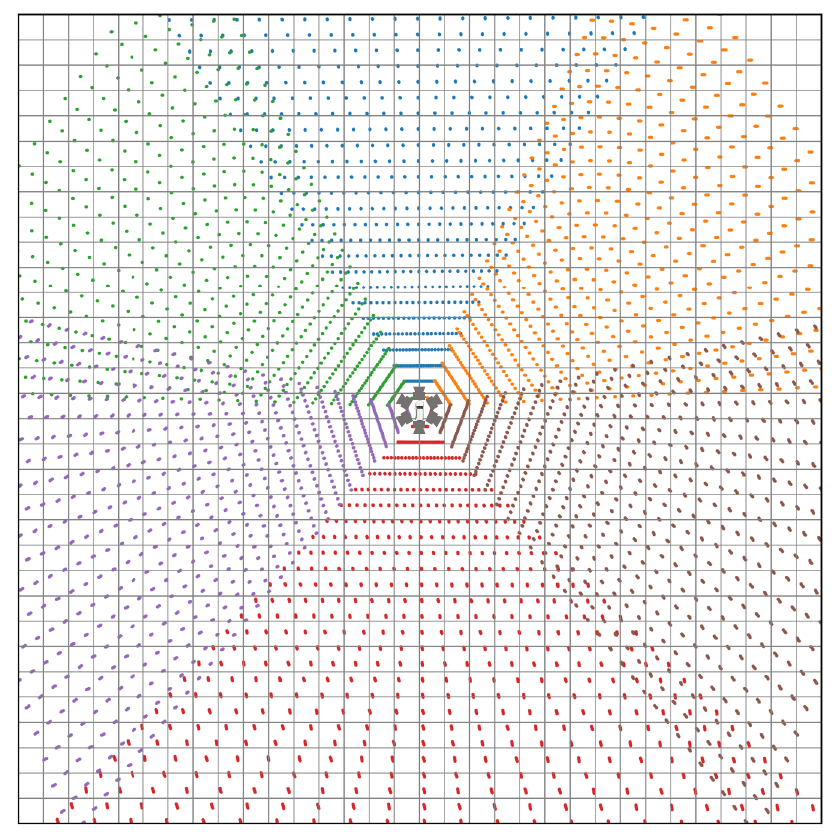}
      \label{fig:cart_grid}
    } 
    \subfloat[Polar BEV]{
      \includegraphics[width=0.45\linewidth]{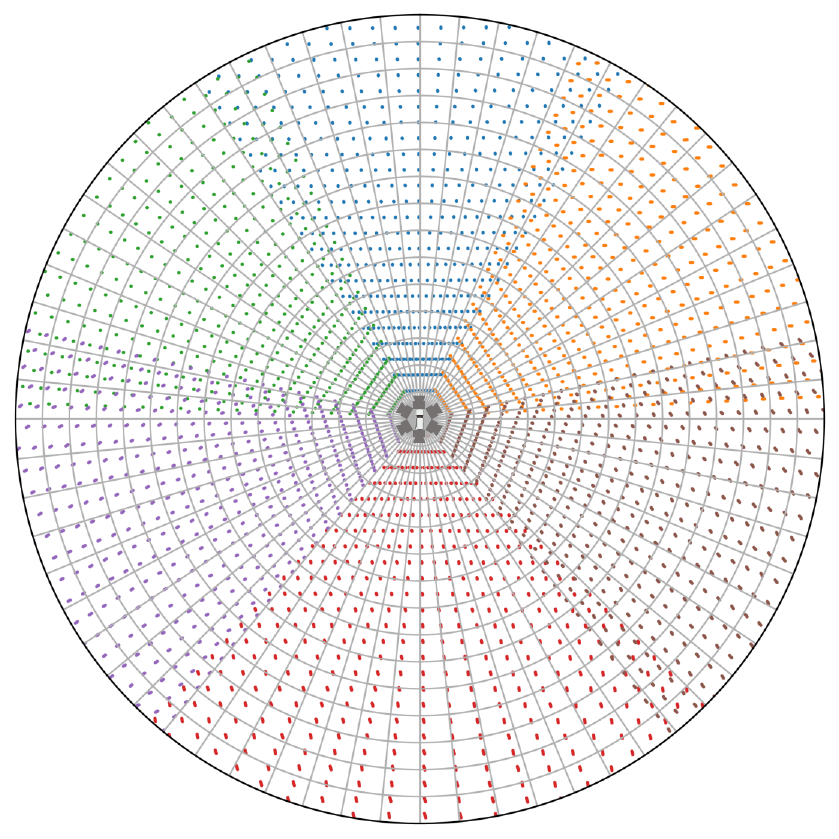}
      \label{fig:polar_grid}
    }
    \caption{
        Illustration of BEV grid distribution and image information distribution. 
        The Cartesian BEV space is uniformly rasterized horizontally and vertically, and the polar BEV 
        space is rasterized angularly and radially. The points with different colors represent the 
        frustum points of the multi-view cameras carrying the image information, and their distribution
        is consistent with the polar grid distribution, which is dense in the near and sparse in the 
        far.  
    }
    \label{fig:grid}
\end{figure}

\begin{figure*}[ht]
  \centering
  \subfloat[Cartesian BEV-based feature extraction and prediction]{
    \includegraphics[width=0.95\linewidth]{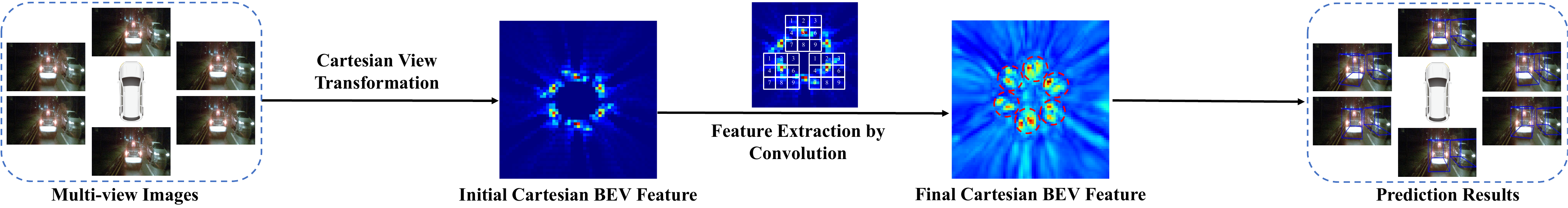}
    \label{fig:cart_view_symmetry}
  } \\
  \subfloat[Polar BEV-based feature extraction and prediction]{
    \includegraphics[width=0.95\linewidth]{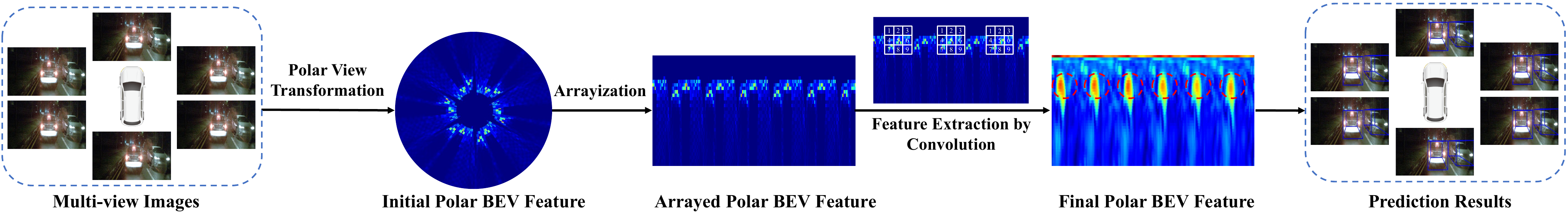}
    \label{fig:polar_view_symmetry}
  }
  \caption{
    Comparison of feature extraction and prediction based on different BEV representations.
      (a). Assuming that the multi-view cameras have the same imaging, the initial Cartesian BEV 
      representation obtained by view transformation is view-symmetric. But the subsequent 
      translation-invariant convolution operation destroys this symmetry, resulting in different 
      object features and predictions at different azimuths.
      (b). When the polar BEV representation is employed, the object features are approximate 
      and parallel in the arrayed polar BEV features. In this way, the view symmetry can be 
      preserved using the regular convolution operation, leading to similar object features and 
      predictions at different azimuths.
  }
  \label{fig:view_symmetry}
\end{figure*}

Recently, many efforts~\cite{huang2021bevdet, huang2022bevdet4d, li2023bevdepth, li2022bevformer, liu2022petr} 
integrate cross-view information with the help of the Bird's-Eye-View~(BEV), which removes inefficient fusion 
post-processing, and achieve significant advances in both detection performance and efficiency. In particular, 
the LSS-based paradigm utilizes the Lift-Splat-Shoot~\cite{philion2020lss} mechanism to construct an 
explicitly dense BEV representation and has become one of the mainstream solutions. It is mainly composed of 
four modules: an image-view encoder for image feature extraction, a view transformer to transform image features 
from image-view to BEV by per-pixel categorical depth estimation, a BEV encoder composed of a series of 2D convolution 
blocks for further BEV feature extraction, and a detection head for 3D object detection in BEV space. In particular, the 
view transformer is pivotal to the whole framework, which bridges the gap between the image coordinate system 
and the Cartesian BEV coordinate system, integrating cross-view information into a unified Cartesian BEV representation. 
The Cartesian BEV representation is intuitive and traditional, but it suffers from the following two 
limitations: \textbf{1). The uniform grid distribution of the Cartesian BEV representation mismatches the 
non-uniform image information distribution.} As shown in Fig.~\ref{fig:grid}(a), the nearby region contains 
richer and denser image information than the distant region, but the Cartesian BEV representation adopts 
the same grid granularity for all different distances. As a result, the grid distribution in the nearby reigon is 
too sparse for the rich information, leading to information loss. Conversely, the grid distribution in the distant 
reigon is too dense for the sparse information, causing computational redundancy. All existing methods overlook this 
limitation because applying the uneven rasterization granularity to the Cartesian BEV space is challenging.
\textbf{2). The Cartesian BEV representation struggles to preserve the view symmetry of surround-view cameras during 
feature extraction.} As shown in Fig.~\ref{fig:view_symmetry}(a), assuming that different cameras obtain the same 
object imaging, the input multi-view images are transformed into view-symmetric BEV features by 
the view transformation. But the subsequent translation-invariant regular 2D convolution operation destroys this symmetry 
during BEV feature extraction, which results in different object features being learned at different azimuths, increasing the 
difficulty of representation learning. AeDet\cite{feng2023aedet} proposes an azimuth-equivariant convolution and an 
azimuth-equivariant anchor to model view symmetry property into the network, but it uses the customized convolution operator, 
which increases the complexity.

In this paper, we innovatively choose the polar BEV representation to replace the Cartesian representation. 
It rasterises the BEV space angularly and radially, eliminating the above limitations at the source: 
\textbf{1). The grid distribution of the polar BEV representation is consistent with the distribution 
of image information, which is dense near and sparse far away as shown in Fig.~\ref{fig:grid}(b).} 
Compared with the Cartesian BEV representation, the polar BEV representation naturally captures 
finer-grained information in the near region, while reducing computational redundancy in the far region. 
\textbf{2). The polar BEV representation can conveniently preserve the view symmetry of surround-view cameras.} 
As shown in Fig.~\ref{fig:view_symmetry}(b), in the polar BEV representation, features corresponding 
to the same object imaging from different views are parallel. Therefore, azimuth-equivalent object features 
can be extracted by simply performing regular 2D convolution operations.

However, switching to the polar BEV representation is non-trivial. To address this, we elaborately tailor three modules: 
a polar view transformer to generate the polar BEV representation, a polar temporal fusion module for fusing cached 
historical polar BEV features and a polar detection head to predict the polar-parameterized representation of the object. 
Specifically, in the polar view transformer, we first lift the multi-view image features to the cylindrical coordinate 
system instead of the traditional Cartesian coordinate system, then the BEV space is rasterized angularly and radially, 
and finally the polar BEV representation is generated by BEV pooling. In the polar temporal 
fusion module, since ego-motion can be more conveniently formulated in the Cartesian coordinate system, we utilize the Cartesian 
coordinate as an intermediate proxy to align the multi-frame polar BEV features. 
In the polar detection head, we adopt polar parameterization for 3D object detection to exploit the preserved view symmetry, 
allowing the azimuth-equivalent prediction targets to be learned. 
Through the cooperation of these three modules, the Cartesian BEV representation is successfully replaced 
with the polar BEV representation. 

In addition, the quality of image features in the perspective view is crucial for detection performance~\cite{yang2023bevformer, Wang2024focal, guo2024cyclic, 2024SongDivide}. 
Many LSS-based works~\cite{li2023bevdepth,li2023bevstereo, park2022solofusion} introduce explicit depth supervision, 
which significantly enhances the depth-awareness of image features, but neglects the object-awareness. 
In this paper, we directly guide the network to learn the object-aware feature representation in the perspective view by 
introducing 2D auxiliary supervision. 
Specifically, we apply object classification supervision in the perspective view to improve the semantic discriminability 
of image features. To enhance the sensitivity of the network to object location, we additionally apply supervision 
on 2D object bounding box and center regression. It is worth noting that these 2D auxiliary tasks are active only 
during training, so they do not slow down inference.

Naturally, the feature extraction in the BEV space is also important. However, it is inevitable that a large amount 
of background noise exists in the BEV representation, which may interfere with the feature extraction for foreground objects. 
To mitigate the negative effects of background noise, we propose a spatial attention enhancement~(SAE) module consisting of two 
convolutional layers. It predicts a spatial attention weight map for weighting with the BEV feature map, guiding the network 
to focus on the foreground region while suppressing the background noise.

We integrate the above improvements to form a novel framework called PolarBEVDet. In summary, the main contributions of this 
paper are as follows:
\begin{itemize}

\item The polar BEV representation is proposed to replace the traditional Cartesian BEV representation for the LSS-based paradigm. 
It can mitigate the information loss in the nearby region and the computational redundancy in the far region, while also conveniently 
preserving the view-symmetry of multiple views. 

\item The 2D auxiliary supervision in perspective view is proposed for the LSS-based paradigm, which enhances the object-awareness of 
image feature representations.

\item The spatial attention enhancement module is proposed to suppress the background noise and highlight the foreground information in 
BEV features, improving the quality of BEV representation.

\item We evaluate the proposed PolarBEVDet on the challenging benchmark nuScenes~\cite{caesar2020nuscenes}. Without any bells and whistles, 
our PolarBEVDet achieves remarkable detection performance~(63.5\% NDS and 55.8\% mAP) on the nuScenes test split.

\end{itemize}

\section{Related Works}
\label{sec:related-work}

\subsection{Multi-view Camera-only 3D Object Detection}
When dealing with multi-camera systems, previous works~\cite{wangfos3d, wang2022pgd, park2021pseudo, Reading2021caddn} utilize 
monocular 3D object detection to process each image separately, and then merge the detection results through 
post-processing in a unified coordinate system. 
This paradigm cannot simultaneously exploit the information from multi-view images and tends to miss truncated 
objects. 
Recent methods transform multi-view features from perspective view to a unified BEV representation for detection, 
which is not only convenient for the integration of multi-view information, but also more conducive to downstream 
tasks such as object tracking and trajectory prediction. In general, these methods can be roughly categorized into 
LSS-based and query-based methods.

\subsubsection{LSS-based methods}
As a pioneer of this paradigm, BEVDet~\cite{huang2021bevdet} and BEVDet4D~\cite{huang2022bevdet4d} follow 
the Lift-Splat-Shoot~\cite{philion2020lss}, which transforms the multi-view features into a dense BEV 
representation by predicting the categorical depth distribution for each pixel in the image feature map. 
BEVDepth~\cite{li2023bevdepth} and BEVStereo~\cite{li2023bevstereo} further introduce explicit depth 
supervision and stereo information to improve the quality of depth estimation. 
BEVPoolv2~\cite{huang2022bevpoolv2} upgrades the view transformation process from the perspective of 
engineering optimization, significantly reducing computation and storage consumption.
SA-BEV~\cite{zhang2023sabev} proposes semantic-aware BEVPooling, which generates semantic-aware BEV 
features by filtering background information based on image semantic segmentation results. 
AeDet~\cite{feng2023aedet} proposes the azimuth-equivariant convolution and azimuth-equivariant
anchor to preserve the radial symmetry properties of BEV features. 
To take full advantage of the temporal information, SOLOFusion~\cite{park2022solofusion} utilizes both short-term, 
high-resolution and long-term, low-resolution temporal stereo.
All of the above methods adopt the Cartesian BEV representation and ignore the object-awareness of image features. 
As far as we know, we are the first to explore the polar BEV representation and exploit 2D auxiliary supervision 
to improve the object-awareness of image features in the LSS-based paradigm.

\subsubsection{Query-based methods}
According to the meaning of query, query-based methods can be categorized into methods based on dense BEV 
query and sparse object query. 

The dense BEV query-based methods generate a dense BEV representation with the help of a set of grid-shaped 
learnable BEV queries. Specifically, BEVFormer~\cite{li2022bevformer}, BEVFormerV2~\cite{yang2023bevformer} 
utilize temporal self-attention and spatial cross-attention to efficiently aggregate spatial-temporal 
information from multi-view images and historical BEV features into BEV queries.

Inspired by Transformer-based 2D object detection~\cite{carion2020detr, zhu2020deformable, liu2022dabdetr}, the sparse object 
query-based methods leverage object queries to represent 3D objects, and utilize the attention mechanism to 
aggregate features directly from multi-view images, thus avoiding explicit view transformation of image features. 
In particular, DETR3D~\cite{wang2022detr3d} projects a set of 3D reference points generated by object queries 
into the multi-view images, and then samples the 2D features of the projection points to update the queries. 
However, this feature querying approach suffers from the problem of inadequate receptive fields.
Therefore, PETR~\cite{liu2022petr} and PETRv2~\cite{liu2023petrv2} propose the 3D position embedding to transform 
image features into 3D position-aware features, and then employ the global attention mechanism for the interaction 
between object queries and image features.
Subsequently, VEDet~\cite{chen2023viewpoint}, CAPE~\cite{xiong2023cape} and 3DPPE~\cite{shu20233dppe} improve the 
3D position embedding of image features and achieve further performance improvement. 
StreamPETR~\cite{wang2023exploring} focuses on long temporal modeling to efficiently propagate 
long-term historical information through object queries, dramatically improving performance at negligible 
computational cost. 
To avoid computationally expensive global attention, SparseBEV~\cite{liu2023sparsebev} designs adaptive 
spatio-temporal sampling to generate sample locations guided by queries, and adaptive mixing to decode sampled 
features with dynamic weights from queries.
While the query-based methods can achieve superior detection performance, they have higher localization errors 
and require greater deployment efforts than the LSS-based paradigm.

\subsection{Polar-Based 3D Perception}
\subsubsection{Polar-based 3D Lidar perception}
Due to the uneven distribution of lidar points in space and even long-tailed distribution, PolarNet~\cite{zhang2020polarnet} 
and Panoptic-PolarNet~\cite{zhou2021panoptic} propose a semantic segmentation and panoptic segmentation network for point 
clouds based on the polar BEV representation, respectively, which balances the number of points in each grid cell and 
indirectly aligns the attention of the segmentation network with the long-tailed distribution of points along the radial axis.
However, these 3D-to-2D projection methods inevitably lose and change the 3D topology. Therefore, Cylinder3D~\cite{Zhu_2021_CVPR} 
and DS-Net~\cite{Hong_2021_CVPR} resort to 3D cylindrical partition and asymmetrical 3D convolution networks for Lidar segmentation 
to better model the geometric information.
When dealing with stream-based lidar perception, PolarStream~\cite{chen2021polarstream} adopts the polar coordinate system 
instead of the Cartesian coordinate system to reduce computation and memory waste.
For Lidar-based 3D object detection, in order to alleviate the feature distortion problem of the polar representation, 
PARTNER~\cite{nie2023partner} designs a global representation re-alignment module consisting of dual attention and 
introduces instance-level geometric information into the detection head, achieving excellent performance beyond the 
Cartesian-based methods. 

\begin{figure*}[ht]
    \centering
    \includegraphics[width=0.9\linewidth]{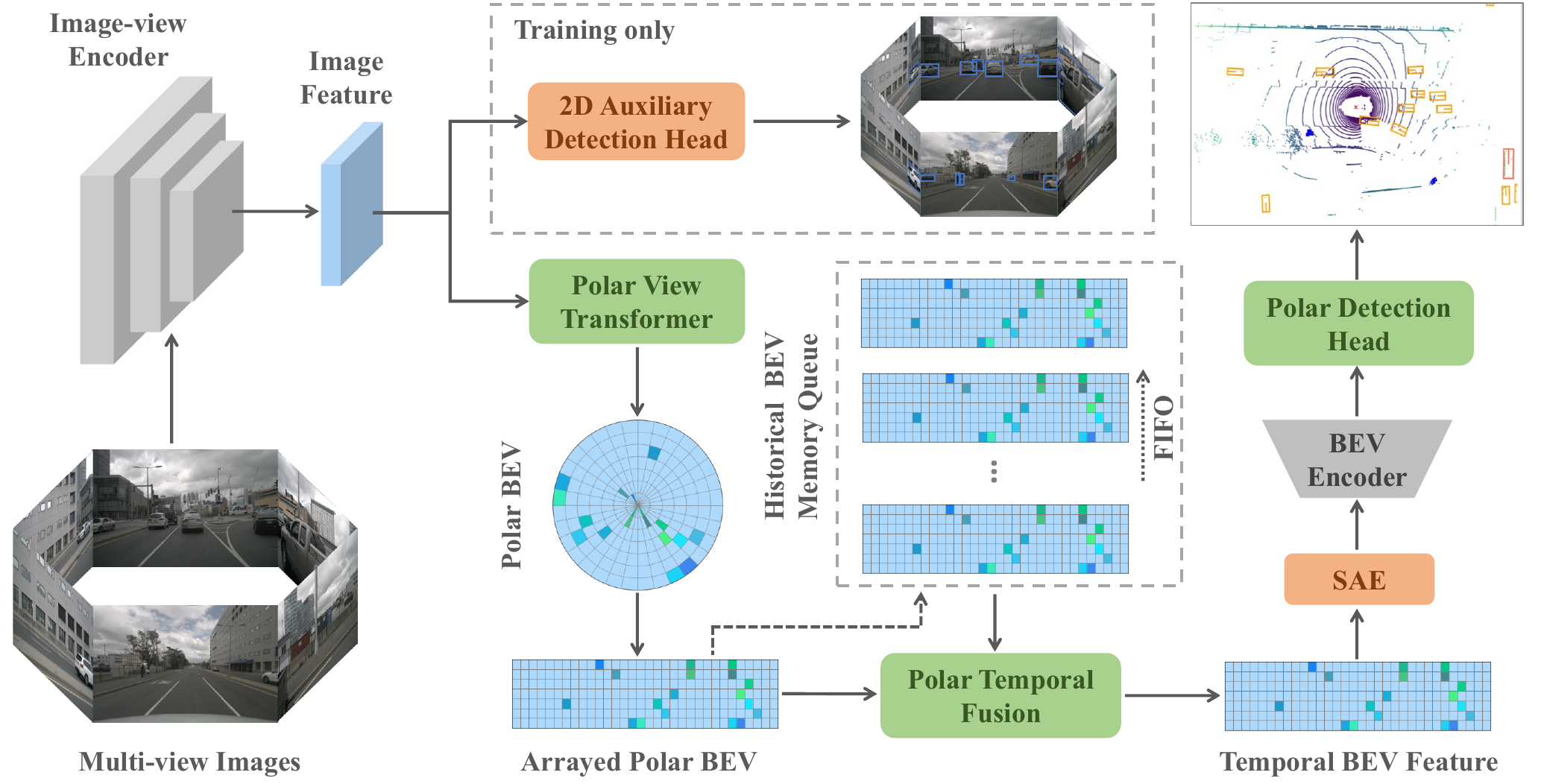}
    \caption{Framework of PolarBEVDet. First, the multi-view image features extracted 
    by the image-view encoder are fed to the polar view transformer to generate a polar 
    BEV representation, which is subsequently arrayed to obtain an arrayed poalr BEV 
    representation. 
    Then, the polar temporal fusion module fuse the cached historical polar BEV features to 
    utilize the temporal information. 
    Finally, the temporal BEV feature is sent to the polar detection head to predict 
    the polar-parameterized representation of the object after further feature extraction 
    by the SAE module and the BEV encoder. 
    In addition, during training phase, the 2D auxiliary detection head is applied to 
    improve the  feature quality in perspective view.}
    \label{fig:framework}
\end{figure*}

\subsubsection{Polar-based 3D vision perception}
PolarBEV~\cite{liu2023vision} employs a height-based feature transformation to generate a polar BEV representation for 
BEV semantic segmentation and instance segmentation, where it adjusts the correspondence between BEV grids and image 
pixels by iteratively estimating the height of each polar grid.
Extending from DETR3D~\cite{wang2022detr3d}, PolarDETR~\cite{chen2022polar} proposes polar parameterization for 3D object 
detection, which establishes explicit associations between image patterns and predicted targets and exploits the view 
symmetry of the surround-view cameras. 
PolarFormer~\cite{jiang2023polarformer} transforms all image feature columns into a set of polar rays associated with 
learnable polar queries through the cross-attention mechanism~\cite{zhu2020deformable}, and these polar rays are subsequently aligned to 
generate multi-scale polar BEV maps for detection. 
These methods all build on the query-based paradigm, in contrast, we focus on the feasibility of polar BEV representation in the LSS-based 
paradigm and explore the capabilities~(e.g., robustness and generalizability) of polar BEV representation in more depth.

% \begin{figure*}[t]
%     \centering
%     \includegraphics[width=0.9\linewidth]{figures/approach/framework1.pdf}
%     \caption{Framework of PolarBEVDet. First, the multi-view image features extracted 
%     by the image-view encoder are fed to the polar view transformer to generate a polar 
%     BEV representation, which is subsequently arrayed to obtain an arrayed poalr BEV 
%     representation. 
%     Then, the polar temporal fusion module fuse the cached historical polar BEV features to 
%     utilize the temporal information. 
%     Finally, the temporal BEV feature is sent to the polar detection head to predict 
%     the polar-parameterized representation of the object after further feature extraction 
%     by the SAE module and the BEV encoder. 
%     In addition, during training phase, the 2D auxiliary detection head is applied to 
%     improve the  feature quality in perspective view.}
%     \label{fig:framework}
% \end{figure*}

\section{Method}
\label{sec:method}
In this paper, we propose a novel multi-view 3D object detection framework, PolarBEVDet, in which we utilize 
the polar BEV representation instead of the Cartesian BEV representation commonly used in LSS-based methods. 
As shown in Fig.~\ref{fig:framework}, we elaborately tailor three modules for the polar BEV representation: a polar view transformer for polar BEV 
representation generation, a polar temporal fusion module for fusing cached historical BEV features and a polar 
detection head for predicting the polar-parameterized representation of the object. 
Furthermore, we impose the 2D auxiliary supervision and the spatial attention enhancement module to improve 
the feature representation quality in perspective view and BEV, respectively.
The image-view encoder and BEV encoder have no special design and simply follow the previous LSS-based 
methods~\cite{huang2021bevdet,li2023bevdepth, park2022solofusion}.

\subsection{Polar View Transformer}
In polar view transformer, we follows the LSS paradigm~\cite{philion2020lss} to transform image features 
$F^{img}=\{F^{img}_k \in \mathbb{R}^{C \times H_F \times W_F}, k=1,2, \dots, N_{view} \}$ into a dense BEV feature for 
subsequent perception. Lift and Splat are two key steps in the LSS paradigm, which are used to lift the image 
features from 2D to 3D space and feature aggregation for generating the BEV feature map, respectively.

In the Lift step, we preset a set of discrete depths $\{d_0, d_1, \cdots, d_{N_D-1}\}$ and utilize a depth estimation 
network to predict a depth distribution $D_k \in \mathbb{R}^{N_D \times H_F \times W_F}$ over these discrete depths 
for each image pixel. The frustum points $P_k \in \mathbb{R}^{3 \times N_D \times H_F \times W_F}$ and corresponding 
frustum features $F_k^{3D} \in \mathbb{R}^{C \times N_D \times H_F \times W_F}$ are derived for each image, where the 
frustum features are obtained by scaling the image features with the predicted depth distribution. To obtain the polar 
BEV representation, we transform all the frustum points of each view into the cylindrical coordinate system instead 
of the Cartesian coordinate system. 
Specifically, given a frustum point $p=(u, v, d)$, it is first projected to a 3D point $p_{cart} =(x, y, z)$ in the  
Cartesian coordinate system according to the camera intrinsic and extrinsic:
\begin{equation}
    [x, y, z, 1]^T= T^{-1}K^{-1}[u*d, v*d, d, 1]^T,
\end{equation}
where $T \in \mathbb{R}^{4 \times 4} $ and $K \in \mathbb{R}^{4 \times 4} $ denote the extrinsic and intrinsic matrices. 
Then, the corresponding cylindrical coordinate $p_{cyli}=(\theta, r, z)$ is computed as follows:
\begin{equation}
    \theta=arctan2(y,x),
\end{equation}

\begin{equation}
    r = \sqrt{{x}^2+{y}^2}.
\end{equation}

In the Splat step, the BEV space is rasterized into a set of polar grids angularly and radially. The azimuth range 
$[-\pi, \pi]$ is uniformly divided into $N_{\theta}$ intervals with an interval of $\delta_{\theta}$, while the radial 
range $[r_{min}, r_{max}]$ is uniformly divided into $N_{r}$ intervals with an interval of $\delta_{r}$. The polar grid 
index $(i, j)$ associated with each frustum point is then computed as follows:
\begin{equation}
    i =\lfloor(\theta+\pi) / \delta_\theta \rfloor,
\end{equation}

\begin{equation}
    j=\lfloor(r - r_{min}) /\delta_r\rfloor.
\end{equation}
The frustum features belonging to the same polar grid are aggregated by sum pooling~\cite{huang2022bevpoolv2} so that 
the 3D frustum features are splatted into a 2D poalr BEV representation. Finally, to facilitate subsequent feature 
extraction via convolutional operations, we array this representation to obtain a regular feature map $F \in \mathbb{R}^{C \times N_{\theta} \times N_{r}}$. 

\subsection{Polar Temporal Fusion}
Temporal information facilitates the detection of the occluded objects and can significantly improve the detection 
performance, especially the velocity prediction~\cite{park2022solofusion, lin2022Sparse4D}. 
In order to efficiently utilize the temporal information, we cache the polar BEV features from previous $N_{t}$ timestamps 
in memory queue following~\cite{park2022solofusion}. When need~ed, we align these historical features to the current timestamp 
for temporal fusion. This temporal alignment process involves determining the position of each element in the current polar 
BEV feature map within the historical feature map based on ego-motion, and then warping the historical features to the current 
coordinate system by bilinear interpolation.

\begin{figure}[t]
    \centering
    \includegraphics[width=\linewidth]{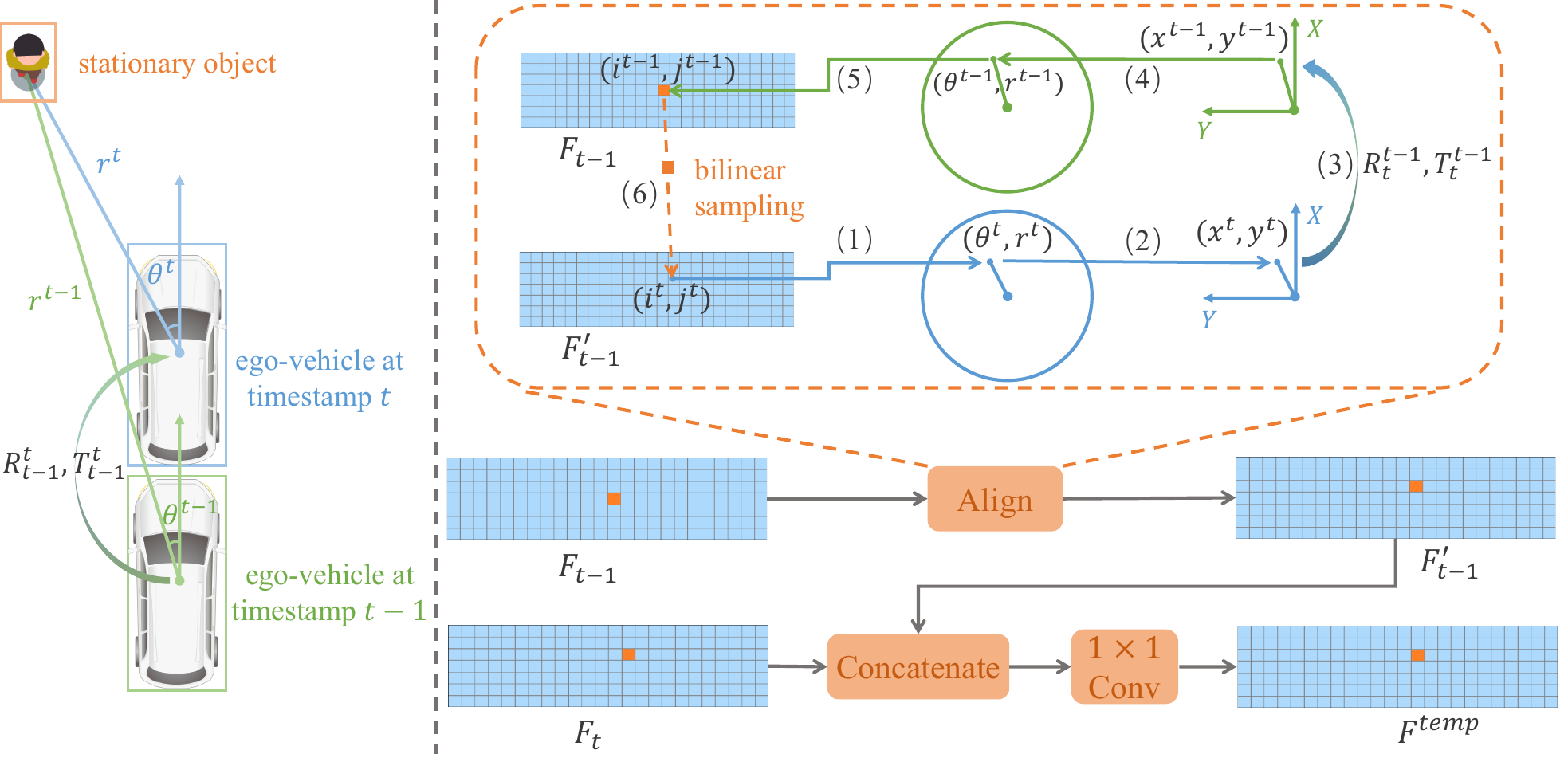}
    \caption{Temporal fusion pipeline for polar BEV feature. The historical feature is aligned according to ego-motion and 
    then fused with the current feature by concatenation and $1 \times 1$ convolution. The details of the temporal alignment 
    is illustrated in orange dashed box. In addition, an example of a scenario is given on the left for ease of understanding.}
    \label{fig:temp_fusion}
\end{figure}

However, describing the ego-motion in the polar coordinate system is challenging, whereas ego-motion can be succinctly and intuitively represented by 
a transformation matrix in the Cartesian coordinate system. Therefore, we use the Cartesian 
coordinates as an intermediate proxy for the alignment of multi-frame polar BEV features. As shown in Fig.~\ref{fig:temp_fusion}, 
given a current polar BEV feature map $F_t \in \mathbb{R}^{C \times N_{\theta} \times N_{r}}$ and a previous polar BEV feature map 
$F_{t-1} \in \mathbb{R}^{C \times N_{\theta} \times N_{r}}$ (for simplicity, we use the case where $N_{t}=1$ as an example), 
we compute the polar coordinate $p^{t}_{p}=(\theta^t, r^t)$ of the element at $(i^t, j^t)$ in the current polar BEV 
feature map:

\begin{equation}
    \theta^t=-\pi+ i^t \times \delta_\theta,
\end{equation}

\begin{equation}
    r^t=r_{min} + j^t \times \delta_r.
\end{equation}
The corresponding Cartesian coordinate $p_{c}^{t}=(x^t, y^t)$ is then computed for simplifying the coordinate transformation:
\begin{equation}
    x^t=r^t \times \cos(\theta^t),
\end{equation}

\begin{equation}
    y^t=r^t \times \sin(\theta^t).
\end{equation}
The ego-motion leads to a position shift of the objects in the feature map, even for stationary objects. Therefore, we 
compensate for this shift by inverse transformation, obtaining the position $p_{c}^{t-1}=(x^{t-1}, y^{t-1})$ of the current 
element in the previous Cartesian coordinate system as follows:

\begin{equation}
    [x^{t-1}, y^{t-1}]^T= R_{t}^{t-1} [x^t, y^t]^T + T_t^{t-1},
\end{equation}
where $R_{t}^{t-1} \in \mathbb{R}^{2\times 2} $ and $T_{t}^{t-1} \in \mathbb{R}^{2}$ represent the horizontal rotation and 
translation matrices of the ego-vehicle from timestamp $t$ to timestamp $t-1$. Then we compute its coordinate 
$p_p^{t-1}=(\theta^{t-1}, r^{t-1})$ in the previous polar coordinate system and its position $(i^{t-1}, j^{t-1})$ in the 
feature map in turn. Finally, the aligned feature map $F_{t-1}'$ at $(i^t, j^t)$ can be obtained by:

\begin{equation}
    F_{t-1}^{'}(i^t, j^t)=bilinear(F_{t-1}, (i^{t-1}, j^{t-1})),
\end{equation}
where $bilinear(F, P)$ denotes the bilinear interpolation of the point $P$ in the feature map $F$, since $(i^{t-1}, j^{t-1})$ may not 
be a valid location in the feature map $F_{t-1}$.

The aligned BEV feature map $F_{t-1}'$ is concatenated with the current BEV feature map $F_{t}$ and then fused by the $1 \times 1$ convolution 
layers to generate the temporal poalr BEV feature $F^{temp} \in \mathbb{R}^{C' \times N_{\theta} \times N_{r}}$.

\begin{figure}[t]
    \centering
    \subfloat[]{
      \includegraphics[width=0.45\linewidth]{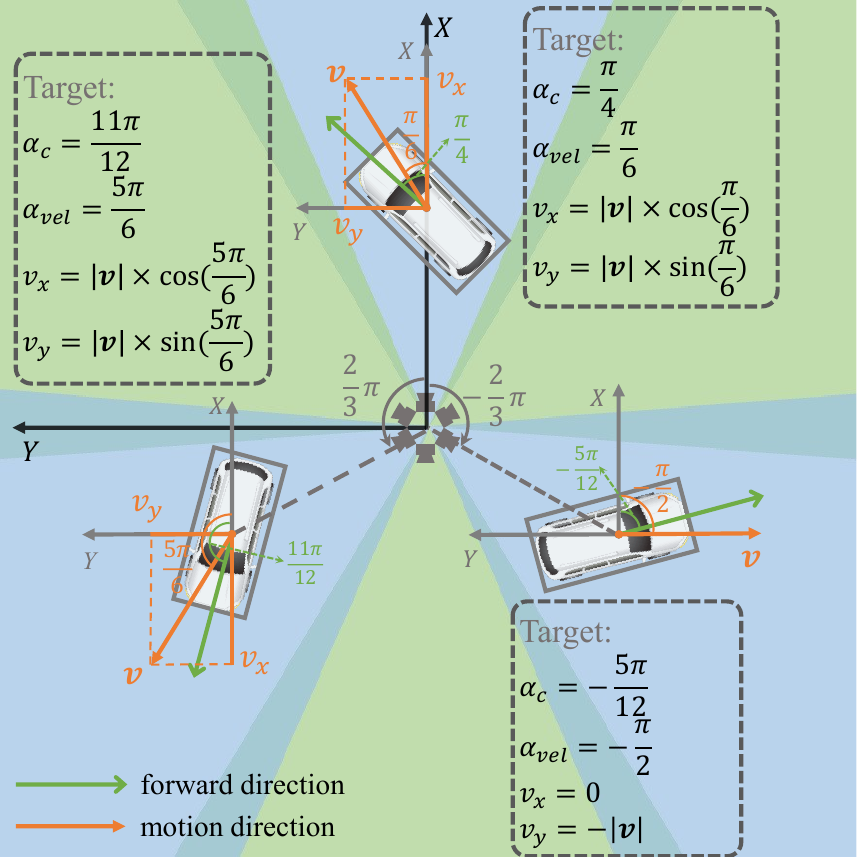}
      \label{fig:cart_view_symmetry}
    } 
    \subfloat[]{
      \includegraphics[width=0.45\linewidth]{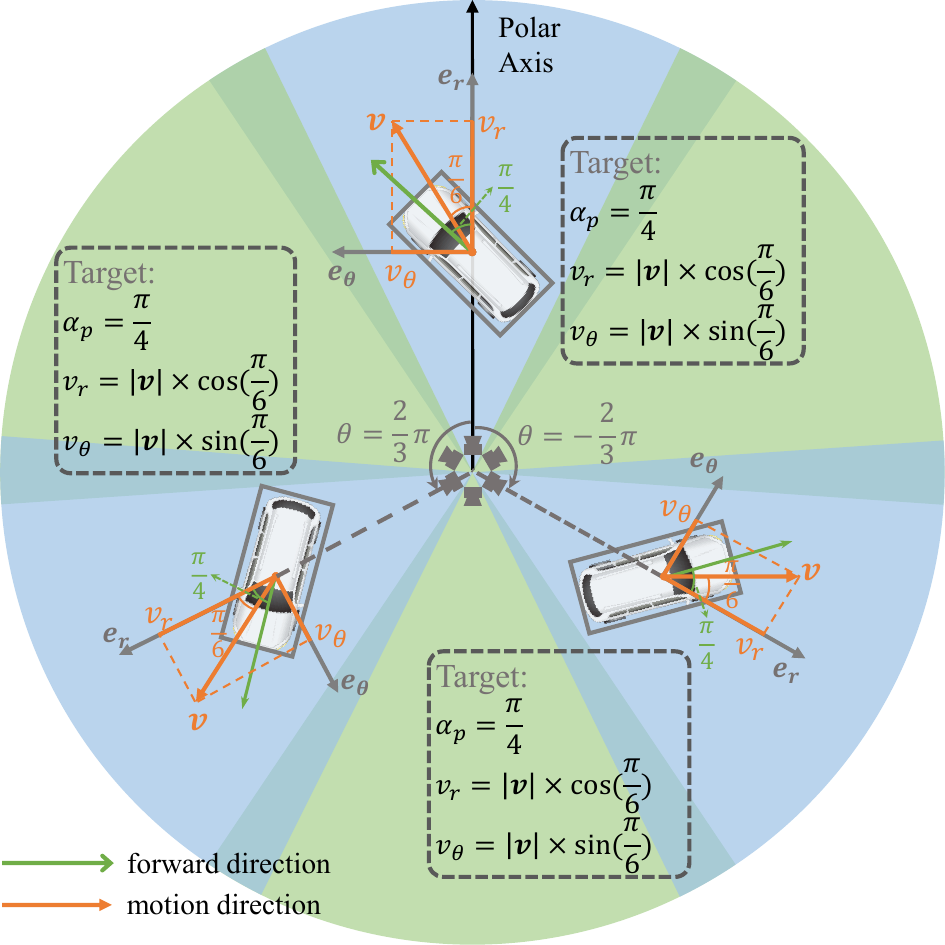}
      \label{fig:polar_view_symmetry}
    }
    \caption{
        Illustration of the Cartesian-parameterized and polar-parameterized prediction targets.
        Assume that multiple identical objects~(represented by white cars) are distributed around 
        the ego-vehicle, and that they are imaged identically in different views. 
        (a) The Cartesian-parameterized prediction targets (for object orientation and velocity) 
        are related to the azimuth of the object, which leads to the same imaging corresponding to 
        different prediction targets.
        (b) In contrast, the polar-parameterized prediction targets are azimuth-equivalent, 
        which reduces the optimization difficulty.
        }
    \label{fig:pred_tartget}
\end{figure}

\subsection{Polar Detection Head}
We apply an anchor-free 3D object detection head following previous works~\cite{Yin2021centerpoint} where the 
predicted target is the polar-parameterized representation of the object rather than the Cartesian-parameterized representation. 

The original detection head contains a center heatmap branch and a group of attribute regression branches. 
The center heatmap branch outputs a 2D heatmap of $K$ channels, each corresponding to one of the $K$ object classes. 
It provides the object class and the rough location $(c_x, c_y)$ of the object’s center in the Cartesian x-y plane.
Meanwhile, the attribute regression branches utilize the center feature to predict the key attributes respectively: 
a sub-voxel center offset $(o_x, o_y)$ to refine the center position, the object’s height $z$,  the 3D dimensions
$(w, l, h)$, the object’s orientation $\alpha_c$ and the velocity $(v_x, v_y)$ along the Cartesian x-y axes.

However, the original detection head is built on the Cartesian BEV representation and parameterization, which is 
not compatible for our polar BEV representation. Moreover, as shown in Fig.~\ref{fig:pred_tartget}(a), for the same 
imaging at different azimuths, it needs to regress different targets for the orientation and velocity of the object, 
inevitably increasing the prediction difficulty~\cite{feng2023aedet}.

Therefore, for the polar BEV representation, we apply the polar parameterization to the object and tailor the polar 
detection head. In particular, the azimuth $\theta$ and radius $r$ are used to describe the location of the object’s center.  
To determine them, we predict a heatmap to locate the rough position $(c_{\theta}, c_r)$ of the object’s center in the
polar coordinate system and a center offset $(o_{\theta}, o_r)$ to compensate for the quantization error. 
For the orientation of the object, the conventional definition is the angle $\alpha_c$ between the object's forward direction 
and the Cartesian x-axis. In order to effectively utilize the view symmetry property preserved by the polar BEV representation, 
we reformulate it as the angle $\alpha_p$ between the forward direction and the polar axis on which the object's center is 
located as follows:
\begin{equation}
    \alpha_p = \alpha_c - \theta.
\end{equation}
Similarly, the original velocity is decomposed into velocity components $(v_x, v_y) $ along the Cartesian x-axis and y-axis, 
respectively. We re-decompose it into radial velocity $v_r$ and tangential velocity $v_\theta$ as prediction targets as follows:

\begin{equation}
    v_r = |\bm{v}| \times \cos(\alpha_{vel}- \theta),
\end{equation}
\begin{equation}
    v_{\theta} = |\bm{v}| \times \sin(\alpha_{vel}- \theta).
\end{equation}
where $|\bm{v}|$ is the magnitude of velocity and $\alpha_{vel}$ is the angle between the motion direction and the Cartesian x-axis. 
For the height and dimensions of the object, the original definition and prediction targets are retained.  
As shown in Fig.~\ref{fig:pred_tartget}(b), the same imaging at different azimuths corresponds to azimuth-equivalent prediction 
targets by polar parameterization, which greatly eases the optimization difficulty. 

As a result, the predicted object is represented as $(\theta=c_{\theta}+o_{\theta}, r=c_r+o_r, z, w, l, h, \alpha_{p}, v_{\theta}, v_{r})$, 
which can be effortlessly converted back to the result of the Cartesian parameterization.

\subsection{2D Auxiliary Detection Head}
In BEV perception system, the quality of image features in perspective view is crucial for the performance of 3D object 
detection~\cite{yang2023bevformer,Wang2024focal}. 
Ideally, image features should be depth-aware and object-aware. However, due to the existence of the view transformer, 
the final object detection in BEV space can only provide implicit and indirect supervision of these awareness capabilities~\cite{yang2023bevformer}.  
This leads to confusing feature representation learning in the perspective view. Many LSS-based works~\cite{li2023bevdepth,li2023bevstereo, park2022solofusion} 
have introduced explicit depth supervision from LiDAR, remarkably improving the depth-awareness of image features, but neglecting the 
object-awareness.

In this paper, we directly guide the network to learn the object-aware feature representation in perspective view by 
introducing auxiliary 2D detection and corresponding supervision as follows:

\subsubsection{Classification}
The classification task aims to differentiate between object classes, including the background, car and pedestrian, etc..
By applying auxiliary classification supervision in the perspective view, image feature representations with higher semantic 
discriminability can be learned. 
Therefore, we attach a lightweight classification branch comprising two convolutional layers to the image features for 
predicting the confidence scores of the $K$ object classes.  During training, hungarian matching~\cite{carion2020detr} is 
adopted for positive/negative sample assignment and the generalized focal loss~\cite{li2020generalized} is used for classification 
supervision.

\subsubsection{Box Regression}
To enhance the network's sensitivity to object position, we introduce an auxiliary supervision for 2D object detection in 
perspective view, which forces the network to learn more discriminative features for localization. These features are 
subsequently transformed into the BEV space to generate BEV representations with high position sensitivity, enabling more 
precise object localization.  
In parallel with the object classification branch, we attach another branch to predict the distance from the pixel position 
to the 4 sides of the bounding box following FCOS~\cite{tian2019fcos}. In the training phase, the 2D ground-truth bounding boxes are 
obtained by projecting the annotated 3D bounding boxes to the image, and L1 loss and GIoU~\cite{rezatofighi2019generalized} 
loss are used for regression supervision.

\subsubsection{Center Regression}
Object detection in BEV space is center-based, so the sensitivity to the object center is very crucial. Inspired by this, 
in order to further improve the precision of object localization, we introduce a center regression branch to predict the 
offset from the pixel position to the object center and supervise it with L1 loss. Besides, we further predict a center 
heatmap for locating the object center and supervise it with penalty-reduced focal loss~\cite{law2018cornernet}. The 
ground-truth center is the projection point of the 3D object's center in the image.

\subsection{Spatial Attention Enhancement}
For 3D object detection task, it is necessary to focus more on the foreground information. However, it is inevitable that 
substantial background noise exists in the BEV feature representations, which may interfere with the feature extraction of 
foreground objects. 
To mitigate the negative impact of background noise, we propose a spatial attention enhancement~(SAE) module. It improves 
the quality of BEV feature representation and detection performance by guiding the network to focus on the foreground regions
while suppressing the background noise.

Given the extracted BEV feature map $F \in \mathbb{R}^{C \times N_{\theta} \times N_{r}}$, an attention weight 
map $M \in \mathbb{R}^{1 \times N_{\theta} \times N_{r}}$ is predicted by feeding it to a subnetwork $\Phi_s$ followed by 
a sigmoid function. 
Its expected values in the foreground and background regions are close to 1 and 0, respectively. By weighting the BEV features 
with it, the features in the foreground region are highlighted and the background noise is suppressed, enabling adaptive feature
enhancement.
Furthermore, to prevent incorrect information suppression and stabilize the training, a skip connection is employed. 
Finally, the enhanced BEV features $F' \in \mathbb{R}^{C \times N_{\theta} \times N_{r}}$ can be represented as:
\begin{equation}
    F'=(1+M)F, ~~ M=sigmoid(\Phi_s(F)),
\end{equation}
where the sub-network $\Phi_s$ consists of a $3 \times 3$ convolutional layer followed by batch normalization and ReLU activation,
and concludes with a $1 \times 1$ convolutional layer for the final output. 
\section{Experiments}

\begin{table*}[t]
    \caption{
        Performance Comparison on the nuScenes Val Split. * Benefited From the Perspective-view Pre-training
    }
    \vspace{-0.05in}
    \label{tab:val}
    \centering
    \small
    
    \begin{tabular}{l|cc|cc|c|cccc}
    \toprule
    \textbf{Method} & \textbf{Backbone} & \textbf{Input Size} & \textbf{NDS$\uparrow$}   & \textbf{mAP$\uparrow$}   & \textbf{mATE$\downarrow$}  & \textbf{mASE$\downarrow$}  & \textbf{mAOE$\downarrow$}  & \textbf{mAVE$\downarrow$}  & \textbf{mAAE$\downarrow$} \\
    \midrule
    BEVDepth~\cite{li2023bevdepth}       & ResNet50 & $256 \times 704$ & 0.475 & 0.351 & 0.639 & \textbf{0.267} & 0.479 & 0.428 & 0.198 \\
    FB-BEV~\cite{li2023fbbev}            & ResNet50 & $256 \times 704$ & 0.498 & 0.378 & 0.620 & 0.273 & 0.444 & 0.374 & 0.200 \\
    AeDet~\cite{feng2023aedet}           & ResNet50 & $256 \times 704$ & 0.501 & 0.387 & 0.598 & 0.276 & 0.461 & 0.392 & 0.196 \\
    BEVPoolv2~\cite{huang2022bevpoolv2}  & ResNet50 & $256 \times 704$ & 0.526 & 0.406 & 0.572 & 0.275 & 0.463 & 0.275 & 0.188 \\
    SOLOFusion~\cite{park2022solofusion} & ResNet50 & $256 \times 704$ & 0.534 & 0.427 & 0.567 & 0.274 & 0.511 & 0.252 & \textbf{0.181} \\
    VideoBEV~\cite{Han2024videobev}      & ResNet50 & $256 \times 704$ & 0.535 & 0.422 & 0.564 & 0.276 & 0.440 & 0.286 & 0.198 \\
    StreamPETR~\cite{wang2023exploring}  & ResNet50 & $256 \times 704$ & 0.540 & 0.432 & 0.581 & 0.272 & 0.413 & 0.295 & 0.195 \\
    SparseBEV~\cite{liu2023sparsebev}    & ResNet50 & $256 \times 704$ & 0.545 & 0.432 & 0.606 & 0.274 & \textbf{0.387} & \textbf{0.251} & 0.186 \\
    \rowcolor{gray!20}
    \textbf{PolarBEVDet} & ResNet50 & $256 \times 704$ & \textbf{0.553} & \textbf{0.450} & \textbf{0.529} & 0.275 & 0.465 & 0.256 & 0.199 \\
    \midrule
    StreamPETR*~\cite{wang2023exploring}  & ResNet50 & $256 \times 704$ & 0.550 & 0.450 & 0.613 & \textbf{0.267} & 0.413 & 0.265 & 0.196 \\
    SparseBEV*~\cite{liu2023sparsebev}    & ResNet50 & $256 \times 704$ & 0.558 & 0.448 & 0.581 & 0.271 & \textbf{0.373} & 0.247 & \textbf{0.190} \\
    \rowcolor{gray!20}
    \textbf{PolarBEVDet*}  & ResNet50 & $256 \times 704$ & \textbf{0.567} & \textbf{0.469} & \textbf{0.525} & 0.269 & 0.437 & \textbf{0.247} & 0.199 \\
    \midrule
    BEVDepth~\cite{li2023bevdepth}        & ResNet101 & $512 \times 1408$ & 0.535 & 0.412 & 0.565 & 0.266 & 0.358 & 0.331 & \textbf{0.190} \\
    AeDet~\cite{feng2023aedet}            & ResNet101 & $512 \times 1408$ & 0.561 & 0.449 & 0.501 & 0.262 & 0.347 & 0.330 & 0.194 \\
    SOLOFusion~\cite{park2022solofusion}  & ResNet101 & $512 \times 1408$ & 0.582 & 0.483 & 0.503 & 0.264 & 0.381 & 0.246 & 0.207 \\
    SparseBEV*~\cite{liu2023sparsebev}    & ResNet101 & $512 \times 1408$ & 0.592 & 0.501 & 0.562 & 0.265 & 0.321 & 0.243 & 0.195 \\
    StreamPETR*~\cite{wang2023exploring}  & ResNet101 & $512 \times 1408$ & 0.592 & 0.504 & 0.569 & \textbf{0.262} & \textbf{0.315} & 0.257 & 0.199 \\
    \rowcolor{gray!20}
    \textbf{PolarBEVDet*} & ResNet101 & $512 \times 1408$ & \textbf{0.602} & \textbf{0.508} & \textbf{0.478} & 0.264 & 0.350 & \textbf{0.231} & 0.203 \\
    \bottomrule
    \end{tabular}
    
    \end{table*}
\subsection{Datasets and Evaluation Metric}
We evaluate our model on the nuScenes~\cite{caesar2020nuscenes} dataset, which is a large-scale autonomous driving 
dataset containing sensor data from six cameras, one LiDAR, and five radars. This dataset consists of 1000 driving 
sequences and it is officially divided into 700/150/150 sequences for training/validation/testing, respectively. 
Each sequence lasts about 20 seconds, with a key sample selected every 0.5 seconds for annotation with 3D bounding boxes.

For 3D object detection, we measure the detection performance using officially provided evaluation metrics, 
including the mean Average Precision~(mAP), nuScenes Detection Score~(NDS), and five true positive~(TP) metrics: 
mean Average Translation Error~(mATE), mean Average Scale Error~(mASE), mean Average Orientation Error~(mAOE), 
mean Average Velocity Error~(mAVE), mean Average Attribute Error~(mAAE). Among them, the mAP metric is used to 
measure the localization precision and it is calculated based on the center distance between the predicted and 
ground-truth objects in the bird's eye view. The NDS is a comprehensive metric that combines mAP with all true 
positive metrics to provide a more holistic performance evaluation for 3D object detection.

\subsection{Implementation Details}
We accomplish the proposed improvement on the reimplemented SOLOFusion~\cite{park2022solofusion}, in which single-task 
detection head replaces multi-task detection head to speed up the inference, and the 3D coordinate system is 
constructed by using the mean optical center of the multi-view cameras as the origin to substitute the lidar/ego coordinate
system. 
During training, the image data augmentation and BEV data augmentation are applied following BEVDet~\cite{huang2021bevdet}. 
PolarBEVDet is trained using the AdamW~\cite{loshchilov2017decoupled} optimizer with weight decay of 1e-7 or 1e-2. 
The learning rate is initialized to 1e-4, and the batch size is set to 32.

When compared with other methods, we perform experiments with ResNet50~\cite{he2016deep}, ResNet101~\cite{he2016deep} and 
V2-99~\cite{lee2019energy} backbones. The ResNet50 and ResNet101 backbones are initialized by ImageNet~\cite{deng2009imagenet} 
and nuImages~\cite{caesar2020nuscenes} pre-training, and the V2-99 backbone is initialized from the DD3D~\cite{park2021pseudo} checkpoint. 
The model is trained for 60 epochs for ResNet50 and ResNet101 without CBGS~\cite{zhu2019class} strategy, and 36 epochs for V2-99 to prevent overfitting. 
For the ablation study, we utilized ResNet50 as the image backbone and the image size is 256$\times$704, and all experiments 
are trained for 24 epochs without using the CBGS~\cite{zhu2019class} strategy. In addition, the default BEV resolution is 256$\times$64 and is 
increased to 384$\times$96 only when a large backbone(i.e. ResNet101, V2-99) is used.

\begin{table*}[t]
    \caption{
        Performance Comparison on the nuScenes Test Split. ConvNeXt-B~\cite{Liu_2022_CVPR} Is Pretrained on 
        ImageNet~\cite{deng2009imagenet}, While V2-99~\cite{lee2019energy} Is Initialized From the DD3D~\cite{park2021pseudo} 
        Checkpoint. The Listed Methods Do Not Use Future Frames During Training or Testing
    }
    \vspace{-0.05in}
    \label{tab:test}
    \centering
    \small
    
    \begin{tabular}{l|cc|cc|c|cccc}
    \toprule
    \textbf{Method} & \textbf{Backbone} & \textbf{Input Size} & \textbf{NDS$\uparrow$}   & \textbf{mAP$\uparrow$}   & \textbf{mATE$\downarrow$}  & \textbf{mASE$\downarrow$}  & \textbf{mAOE$\downarrow$}  & \textbf{mAVE$\downarrow$}  & \textbf{mAAE$\downarrow$} \\
    \midrule
    BEVDepth~\cite{li2023bevdepth}        & V2-99 & $640\times1600$ & 0.600 & 0.503 & 0.445 & 0.245 & 0.378 & 0.320 & 0.126 \\
    BEVStereo~\cite{li2023bevstereo}      & V2-99 & $640\times1600$ & 0.610 & 0.525 & 0.431 & 0.246 & 0.358 & 0.357 & 0.138 \\
    AeDet~\cite{feng2023aedet}            & ConvNeXt-B & $640\times1600$ & 0.620 & 0.531 & 0.439 & 0.247 & 0.344 & 0.292 & 0.130 \\
    SOLOFusion~\cite{park2022solofusion}  & ConvNeXt-B & $640\times1600$ & 0.619 & 0.540 & 0.453 & 0.257 & 0.376 & 0.276 & 0.148 \\
    SA-BEV~\cite{zhang2023sabev}          & V2-99 & $640\times1600$ & 0.624 & 0.533 & 0.430 & \textbf{0.241} & 0.338 & 0.282 & 0.139 \\
    FB-BEV~\cite{li2023fbbev}             & V2-99 & $640\times1600$ & 0.624 & 0.537 & 0.439 & 0.250 & 0.358 & 0.270 & 0.128 \\
    SparseBEV~\cite{liu2023sparsebev}     & V2-99 & $640\times1600$  & 0.627 & 0.543 & 0.502 & 0.244 & \textbf{0.324} & 0.251 & \textbf{0.126} \\
    VideoBEV~\cite{Han2024videobev}       & ConvNeXt-B & $640\times1600$ & 0.629 & 0.554 & 0.457 & 0.249 & 0.381 & 0.266 & 0.132 \\
    \rowcolor{gray!20}
    \textbf{PolarBEVDet} & V2-99 & $640 \times 1600$ & \textbf{0.635} & \textbf{0.558} & \textbf{0.429} & 0.253 & 0.389 & \textbf{0.247} & 0.127 \\
    \bottomrule
    \end{tabular}
    
    \end{table*}

\subsection{Main Results}
\subsubsection{nuScenes val split}
In Tab.~\ref{tab:val}, we compare PolarBEVDet with previous state-of-the-art vision-based multi-view 3D detectors 
on the nuScenes validation split. When adopting ResNet50~\cite{he2016deep} as the image backbone and the input size is 256$\times$704,
PolarBEVDet outperforms the previous state-of-the-art LSS-based detector SOLOFusion~\cite{park2022solofusion} by a 
clear margin of 1.9\% mAP and 2.1\% NDS, and also outperforms the query-based detector SparseBEV~\cite{liu2023sparsebev} 
by 1.4\% mAP and 1.0\% NDS. 
By equipping the nuImages~\cite{caesar2020nuscenes} pre-training, PolarBEVDet achieves a state-of-the-art performance of 
56.7\% NDS, which demonstrates its superior detection capability.
Furthermore, by adopting ResNet101~\cite{he2016deep} as the image backbone and scaling up the input size to 512$\times$1408, 
the performance of PolarBEVDet is remarkably improved and still outperforms other previous state-of-the-art detectors, which 
demonstrates the scalability of our proposed method for backbones with different capacities and varying input sizes.

\subsubsection{nuScenes test split}
In Tab.~\ref{tab:test}, we report the performance comparison results on the test split, and it can be observed that PolarBEVDet 
achieves excellent detection performance. Notably, PolarBEVDet outperforms SOLOFusion~\cite{park2022solofusion} with ConvNext-Base~\cite{Liu_2022_CVPR} 
backbone by 1.6\% NDS and 1.8\% mAP, even though we adopt a smaller V2-99~\cite{lee2019energy} backbone. 
In addition, compared to the query-based detector SparseBEV~\cite{liu2023sparsebev}, PolarBEVDet significantly outperforms 
it by 1.5\% mAP and 0.8\% NDS and achieves 7.3\% remarkably less mATE, demonstrating the effectiveness of our method and 
its superiority in object localization precision.

\subsection{Ablation Studies}

\subsubsection{Component analysis}
To analyze the impact of each component in PolarBEVDet, we report the overall ablation study in Tab.~\ref{tab:abl_component}. 
We take the reproduced SOLOFusion as the baseline, and then gradually add the polar representation, 2D auxiliary supervision, 
and the spatial feature enhancement module. By replacing the Cartesian BEV representation with the polar representation, the performance 
is boosted by 1.7\% and 1.3\% on mAP and NDS, respectively. Subsequently, by introducing 2D auxiliary supervision 
during training, a 1.5\% mAP and 1.2\% NDS improvement is observed without increasing the latency. 
Finally, we incorporate the spatial feature enhancement to emphasize the foreground information and suppress the noise, 
resulting in 0.5\% mAP and 0.6\% NDS improvement.  

\begin{table}[t]
    \centering
    \caption{
        Ablation of different components in PolarBEVDet. The baseline is SOLOFusion~\cite{park2022solofusion} with an input resolution of 
        $256\times704$, ResNet50 as the backbone, and a long-term history of 16 frames
        }
    \label{tab:abl_component}

    \resizebox{\columnwidth}{!}{
        \begin{tabular}{cccc|cc}
        \toprule
        \textbf{Baseline} & \makecell[c]{\textbf{Polar}\\ \textbf{Representation}} & \makecell[c]{\textbf{2D Auxiliary}\\ \textbf{Supervision}} & \textbf{SAE} & \textbf{NDS$\uparrow$} & \textbf{mAP$\uparrow$}\\
        \midrule
        \checkmark & - & - & - & 0.496 & 0.397 \\
        \checkmark & \checkmark & - & - & 0.509 & 0.414 \\
        \checkmark & \checkmark & \checkmark & - & 0.521 & 0.429 \\
        \checkmark & \checkmark & \checkmark & \checkmark & \textbf{0.527} & \textbf{0.434}\\
        
        \bottomrule
        \end{tabular}
    }
    \end{table}

\begin{table*}[t]
    \centering
	\caption{Performance of Cartesian and Polar Representations in the Original View and Revolved View}
    \label{tab:revolve_test}
    {
        \begin{tabular}{c@{\hspace{1.5\tabcolsep}}c@{\hspace{1.5\tabcolsep}}cccccccc}
            \toprule
            \textbf{Representation } & \textbf{Original view} & \textbf{Revolved view} & \textbf{NDS}$\uparrow$ & \textbf{mAP}$\uparrow$  & \textbf{mATE}$\downarrow$ & \textbf{mASE}$\downarrow$  & \textbf{mAOE}$\downarrow$ & \textbf{mAVE}$\downarrow$ & \textbf{mAAE}$\downarrow$ \\
            \cmidrule(r){1-1}
            \cmidrule(r){2-3}
            \cmidrule(r){4-10}
            Cartesian & \checkmark &            & 0.496 & 0.397 & 0.613 & 0.295 & 0.660 & 0.265 & 0.192 \\
                      &            & \checkmark & 0.472 & 0.383 & 0.643 & 0.300 & 0.729 & 0.323 & 0.198 \\
            \cmidrule(r){1-1}
            \cmidrule(r){2-3}
            \cmidrule(r){4-10}
            Polar     & \checkmark &            & 0.509 & 0.414 & 0.579 & 0.290 & 0.610 & 0.296 & 0.209 \\
                      &            & \checkmark & 0.505 & 0.411 & 0.588 & 0.291 & 0.624 & 0.292 & 0.212 \\
            \bottomrule
        \end{tabular}
    }
\end{table*}

\begin{figure}[h]%
  \centering
  \subfloat[]{
    \includegraphics[width=0.8\linewidth]{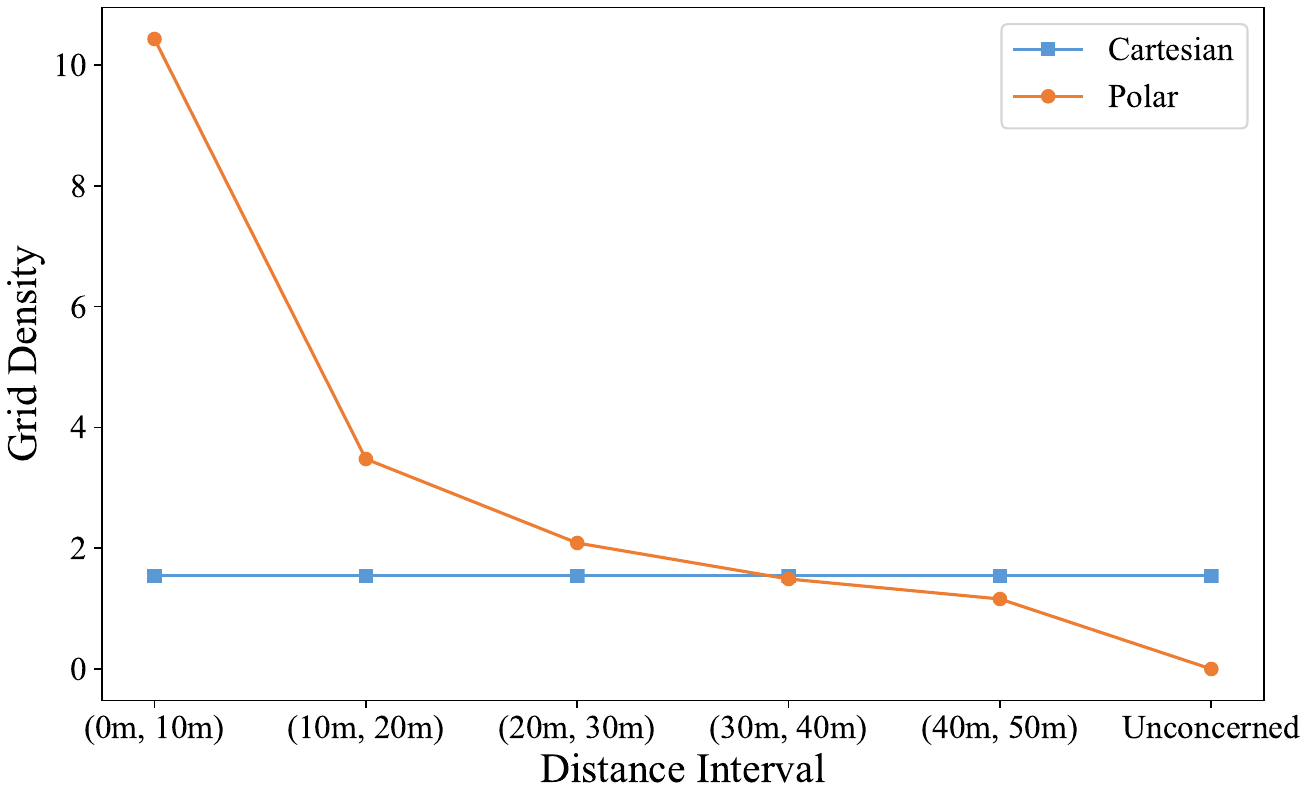}
    \label{fig:density_vs_dis}
    }\hfill
  \subfloat[]{
    \includegraphics[width=0.8\linewidth]{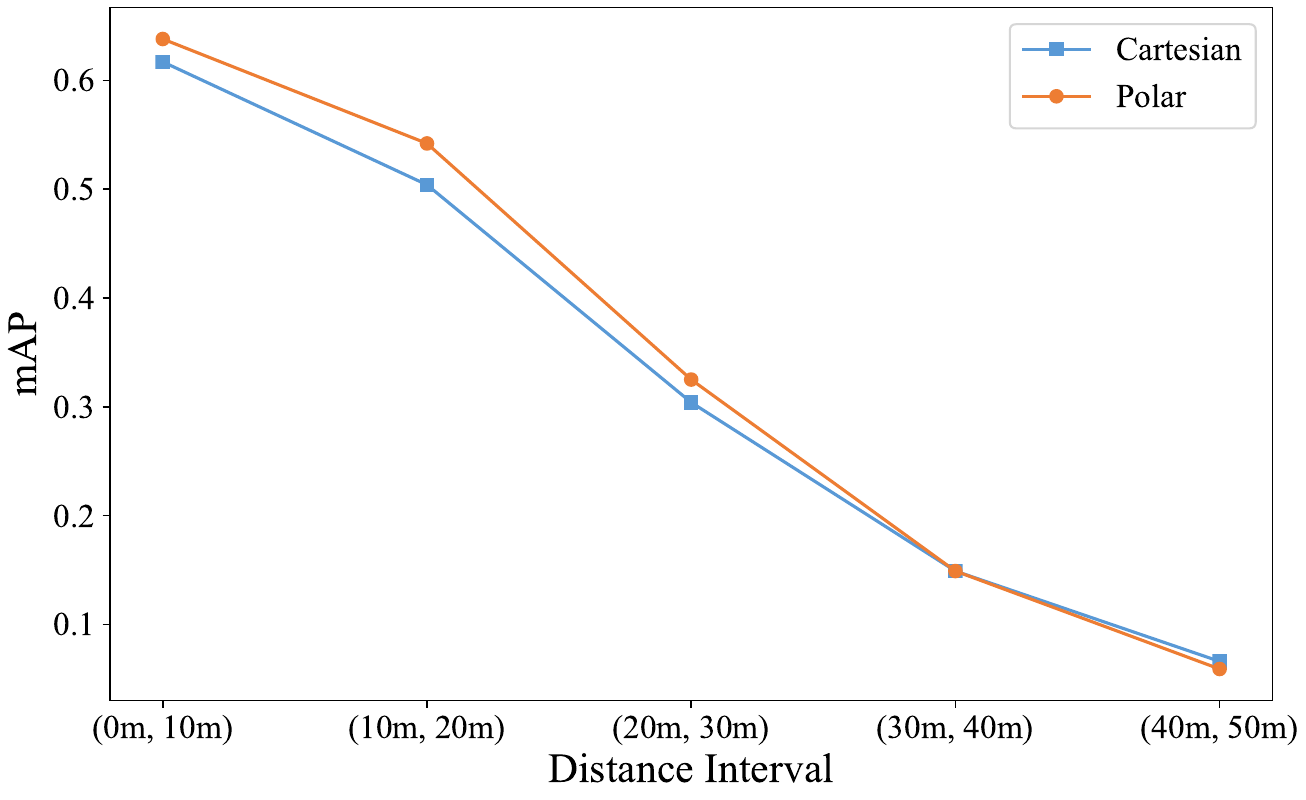}
    \label{fig:map_vs_dis}
    }\\    
  \caption{
    Comparison of grid density and detection performance between Cartesian and polar BEV representations 
    in different distance intervals.
    (a) The grid of Cartesian BEV representation is uniformly distributed, even in unconcerned regions. 
    In contrast, the polar BEV representation has a grid distribution that is dense in the near and sparse 
    in the far, and it can ignore the unconcerned regions.
    (b) The detection performance of the polar BEV representation is significantly better than that of the 
    Cartesian BEV representation in the near region, while it is only slightly degraded in the far region.
  }
  \label{fig:dis_performance}
\end{figure}

\subsubsection{Polar vs. Cartesian at different distances}
The Cartesian BEV and the polar BEV representation possess distinctly different grid distribution.  In Fig.~\ref{fig:dis_performance}(a), 
we count the grid density corresponding to different distance intervals, which is defined as the number of grids contained 
per unit area. Notably, the Cartesian BEV representation has a resolution of $128 \times 128$ while the polar BEV representation has a 
resolution of $256 \times 64$, so the comparison is fair with the same amount of computation. 
It can be readily observed that the Cartesian representation has a uniform grid distribution, while the polar 
representation has a dense near and sparse far distribution. This suggests that with the polar BEV representation, 
the network's attention is more focused on the near area. This is close to the way human drivers perceive, since human 
drivers also tend to focus and react more on things and situations in the near area.

In addition, we compare the perception performance of Cartesian and polar representations for different distance intervals. 
As shown in Fig.~\ref{fig:dis_performance}(b), in the range of $0-30$ m, the polar BEV representation has a significant improvement 
on mAP, which indicates that the polar representation is more advantageous for the perception in nearby regions.  While in the range of $40-50$ m, 
there is a slight decrease in the detection performance of the polar BEV representation, which is due to the sparse grid distribution 
in the distant regions. Nevertheless, this decline is acceptable because the polar representation significantly optimizes the perception 
of critical nearby areas, resulting in overall benefits and better aligning with the actual perception needs of driving.

\subsubsection{Polar vs. Cartesian on robustness to different azimuths}
When a vehicle turns at a large angle, the camera orientation changes significantly. To accurately detect surrounding objects 
under these conditions, the autonomous driving system must be robust to changes in azimuth. To reveal the effect of polar and 
Cartesian representations on the detection robustness to different azimuths, we follow AeDet~\cite{feng2023aedet} to perform the revolving 
test. Specifically, we rotate the vehicle 60 degrees clockwise and the original view ['FRONT LEFT', 'FRONT', 'FRONT RIGHT', 'BACK RIGHT', 'BACK', 'BACK LEFT'] 
becomes the revolved view ['BACK LEFT', 'FRONT LEFT', 'FRONT', 'FRONT RIGHT', 'BACK RIGHT', 'BACK'], and then we evaluate the detection 
performance of polar- and Cartesian-based detectors in the original view and the revolved view, respectively.

As shown in Tab.~\ref{tab:revolve_test}, for the Cartesian BEV representation, replacing the original view with the revolved view decreases 
the mAP and NDS by 1.4\% and 2.4\%, respectively, and the translation, orientation, and velocity errors also grow significantly. This suggests 
that the azimuth change has a significant adverse effect on the performance of the Cartesian-based detector. On the contrary, for the polar BEV 
representation, the detection performance in the rotated viewpoint is only slightly degraded, with mAP and NDS decreasing by only 0.3\% and 0.4\%. 
This suggests that the polar-based detector is more robust to azimuth changes, due to the fact that its feature learning and prediction targets 
are azimuth-independent.

\begin{table}[h]
    \centering
	\caption{Comparision of LSS-based Baselines and Corresponding Polar Versions. The Polar Versions 
    Replace the Cartesian BEV Representation in Baselines with the Polar BEV Representation}
    \label{tab:generalization}
    {
        \resizebox{\linewidth}{!}{
        \setlength{\tabcolsep}{1.5pt}
        \begin{tabular}{l|cc|ccccc}
        \toprule
        \textbf{Method} & \textbf{NDS$\uparrow$}  & \textbf{mAP$\uparrow$}   & \textbf{mATE$\downarrow$}  & \textbf{mASE$\downarrow$}  & \textbf{mAOE$\downarrow$}  & \textbf{mAVE$\downarrow$}  & \textbf{mAAE$\downarrow$} \\
        \midrule
        BEVDet~\cite{huang2021bevdet} & 0.360 & 0.279 & 0.774 & 0.286 & 0.659 & 0.827 & 0.247 \\
        \rowcolor{gray!20}
        +Polar                        & 0.382 & 0.302 & 0.739 & 0.278 & 0.558 & 0.877 & 0.238 \\
        \midrule
        BEVDet4D~\cite{huang2022bevdet4d} & 0.426 & 0.299 & 0.767 & 0.285 & 0.608 & 0.374 & 0.198 \\
        \rowcolor{gray!20}
        +Polar                            & 0.450 & 0.323 & 0.714 & 0.289 & 0.559 & 0.360 & 0.190 \\
        \midrule
        BEVDepth~\cite{li2023bevdepth}    & 0.457 & 0.338 & 0.670 & 0.281 & 0.580 & 0.360 & 0.225 \\
        \rowcolor{gray!20}
        +Polar                            & 0.474 & 0.352 & 0.656 & 0.282 & 0.526 & 0.369 & 0.193 \\
        \bottomrule
        \end{tabular}
        }
    }
\end{table}

\subsubsection{Generalization of the polar representation}
To verify the generalization and portability of the polar BEV representation, we port it to multiple LSS-based detectors and then 
evaluate the performance gain it brings. During porting, we simply replace the original Cartesian-based modules with our designed 
polar view transformer, polar temporal fusion module and polar detection head for fair comparison.

We select BEVDet~\cite{huang2021bevdet}, BEVDet4D~\cite{huang2022bevdet4d} and BEVDepth~\cite{li2023bevdepth} as baselines and 
compare them with their corresponding polar versions as shown in Tab.~\ref{tab:generalization}. The results show that both mAP 
and NDS of the polar versions obtain consistent improvements compared to the baselines, and the translation and orientation 
errors are significantly degraded. This demonstrates that the polar BEV representation has superior generalization ability 
across different models.

\begin{figure*}[t]
  \centering
  \includegraphics[width=\linewidth]{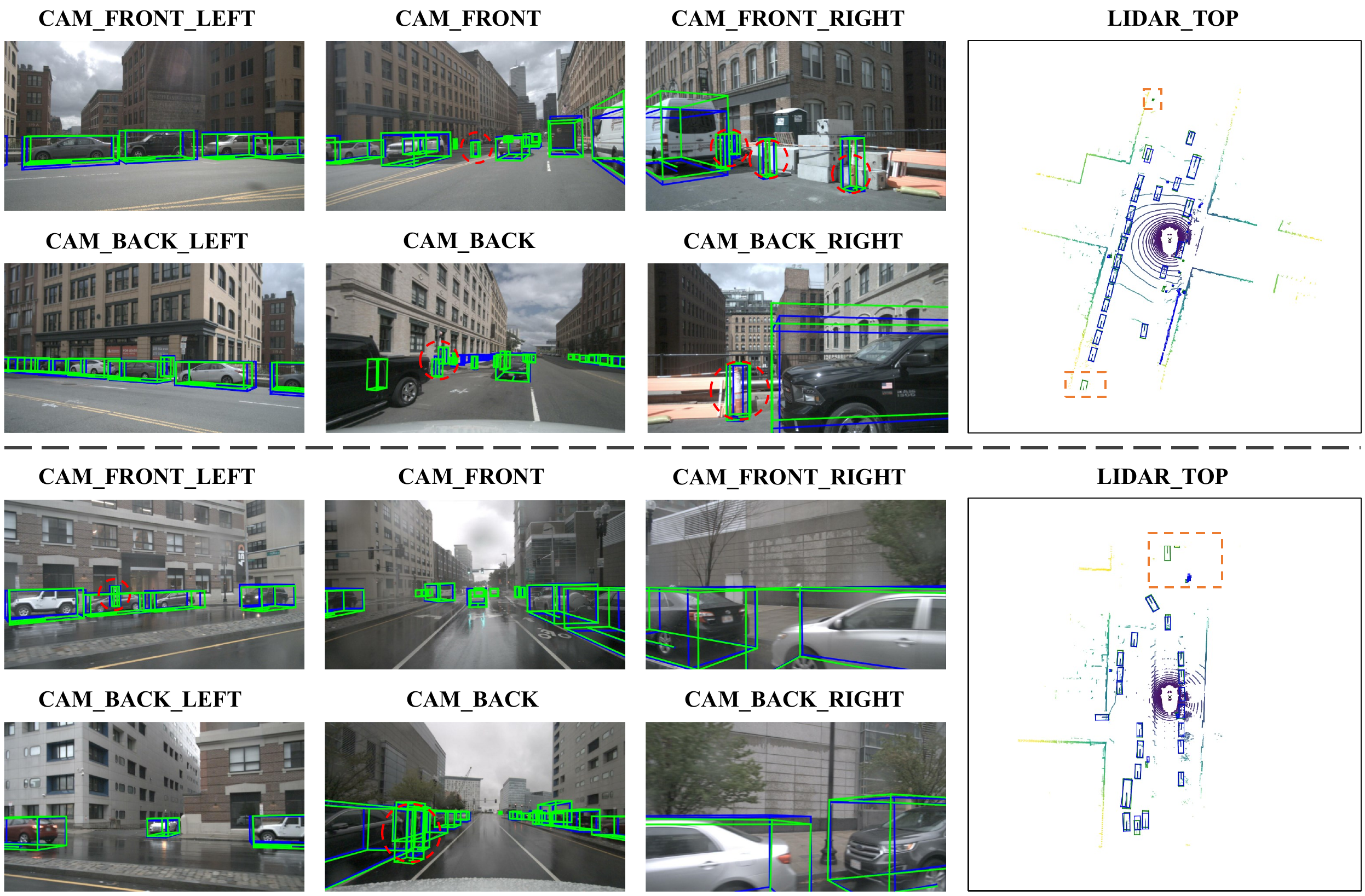}
  \caption{Qualitative results of prediction results in perspective view and BEV. 
  The predicted and ground-truth objects are denoted by blue and green boxes, respectively.
  In addition, some success cases of small object detection and failure cases of distant 
  object detection are marked by red circles and orange rectangular boxes, respectively.}
  \label{fig:vis}
\end{figure*}

\subsubsection{Effect of 2D auxiliary supervision}
To further explore the role of each sub-task in 2D auxiliary supervision, we conduct an ablation study, as shown in Tab.~\ref{tab:2dsup}. 
Starting with the baseline without 2D auxiliary supervision, the initial integration of box regression supervision to improve sensitivity 
to object’s location results in a 0.6\% and 0.7\% increase in mAP and NDS, respectively. Subsequently, we introduce classification 
supervision to improve the semantic discriminability of features, further increasing mAP by 0.5\% and NDS by 0.4\%. Finally, 
by incorporating center regression supervision to improve the sensitivity to the object’s center, our method achieves a 0.4\% mAP and 
0.2\% NDS improvement. Overall, these 2D auxiliary supervision tasks can effectively improve the detection performance by enhancing 
the quality of image features.
\begin{table}[h]
    \centering
	\caption{Ablation of 2D Auxiliary Supervision. The Supervision of Box Regression, 
    Classification, and Central Regression Is Added Sequentially}
    \label{tab:2dsup}
    {
        \begin{tabular}{l|cc}
            \toprule
            \textbf{Setting} & \textbf{NDS$\uparrow$} & \textbf{mAP$\uparrow$} \\ 
            \midrule
            w/o 2D Supervision  & 0.514 & 0.419  \\
            + Box Regression    & 0.521 & 0.425  \\
            + Classification    & 0.525 & 0.430  \\
            + Center Regression & \textbf{0.527} & \textbf{0.434}  \\
            \bottomrule
        \end{tabular}
    }
\end{table} 

\subsubsection{Effect of insertion position of SAE}
The optimal insertion location of the SAE module is difficult to determine directly, so we address this problem by comparing some potential 
insertion locations in Tab.~\ref{tab:sae}. The results show that the detection performance is sensitive to the insertion location of the SAE 
module. Among all configurations, inserting the SAE module before the BEV encoder in configuration C is the optimal choice, 
achieving the highest 43.4\% mAP and 52.1\% NDS. In configuration B, when the SAE module is inserted before temporal fusion, 
mAP and NDS decrease by 0.5\% and 0.4\% compared to configuration A without SAE. This indicates that SAE negatively affects 
the temporal fusion since it may lead to feature inconsistency of the object across frames. When we postpone the SAE module to the back of the 
BEV encoder in configuration D, mAP and NDS decrease by 0.4\% and 0.2\% compared to configuration C. 
This indicates that employing SAE before the BEV encoder to suppress background noise and emphasize foreground information is more conducive 
to BEV feature extraction.
\begin{table}[h]
    \centering
	\caption{Performance Comparison of Different Insertion Positions of SAE}
    \label{tab:sae}
    {
        \begin{tabular}{c|l|cc}
            \toprule
            \textbf{Configuration} & \textbf{Instertion Position} & \textbf{NDS$\uparrow$} & \textbf{mAP$\uparrow$} \\ 
            \midrule
            A & None                    & 0.521          & 0.429           \\
            B & Before Temporal Fusion  & 0.517          & 0.424           \\
            C & Before BEV Encoder      & \textbf{0.527} & \textbf{0.434}  \\
            D & After BEV Encoder       & 0.525          & 0.430           \\
            \bottomrule
        \end{tabular}
    }
\end{table}

\subsection{Qualitative Results}
The qualitative detection results are shown in Fig.~\ref{fig:vis}. The 3D bounding boxes are 
projected and drawn in BEV space and perspective views. We can observe that the predicted bounding 
boxes are close to the ground-truths even for small objects (e.g.~traffic cones and pedestrians), 
proving the superior detection performance of PolarBEVDet. However, our method still has some missed 
and duplicate detections for remote objects, which is a common and challenging issue for camera-based methods.

\section{Conclusion}
In this paper, we propose a novel LSS-based multi-view 3D object detection framework, PolarBEVDet. 
It transforms the multi-view image features into a polar BEV representation instead of a Cartesian BEV 
representation, which is able to better adapt the image information distribution and can easily preserve 
the view symmetry. In addition, it also improves the quality of feature extraction by applying 2D auxiliary 
supervision and spatial attention enhancement in perspective view and BEV, respectively. Extensive experiments 
validate the effectiveness of these improvements and show that PolarBEVDet achieves remarkable performance. 
Notably, the polar BEV representation has significant advantages in both near-range perception and robustness 
of azimuth changes, and porting experiments have also demonstrated its superior universality. Therefore, 
the polar BEV representation, as a comparable or even superior alternative to the Cartesian BEV representation, 
holds promise in further motivating future research on the LSS paradigm. 

\label{sec:conclusion}

\bibliographystyle{IEEEtran}
\bibliography{PolarBEVDet}

\end{document}